\documentclass[runningheads]{llncs}
\usepackage{amsmath,amssymb}
\usepackage{graphicx}
\usepackage{array}
\usepackage{booktabs}
\usepackage{xcolor}
\usepackage{placeins}
\usepackage{cite} 
\usepackage[hidelinks]{hyperref}

\begin{document}

\title{Sharpening the Ensemble: An SSIM-Aligned Residual Refiner\\for Brain-MRI Inpainting Post-Processing}
\titlerunning{Sharpening the Ensemble}

\author{Kubilay Ka\u{g}an K\"om\"urc\"u \and \.Ilkay \"Oks\"uz}
\authorrunning{K. K. K\"om\"urc\"u and \.I. \"Oks\"uz}
\institute{Department of Computer Engineering, Istanbul Technical University,\\
Istanbul, T\"urkiye\\
\email{komurcu17@itu.edu.tr, oksuzilkay@itu.edu.tr}}

\maketitle

\begin{abstract}
Brain-MRI inpainting replaces a masked region of a scan with synthesized,
anatomically plausible healthy tissue, so that analysis tools built for healthy brains can
be applied to images they would otherwise reject. On the BraTS local-synthesis benchmark,
which ranks submissions on the structural similarity index (SSIM), the peak signal-to-noise
ratio, and the mean squared error (MSE) jointly, the strongest recent models are accurate,
but several report blurry synthesized regions and attribute this to the mean-seeking
behavior of the $\ell_1$ and MSE terms in their training losses. We address this in
post-processing, forming a deep ensemble of the two co-first-place 2025 models and training a
lightweight residual refiner on the ensemble's own outputs under an $\ell_1$ loss augmented
with a structural-similarity term whose weight $\lambda$ we vary. At a moderate $\lambda$ the
refiner improves SSIM over the ensemble, from $0.8767$ to $0.8780$ on a held-out reproduction
of the official scorer and from $0.8555$ to $0.8572$ on the official validation leaderboard,
with essentially no change in MSE. The gain is small but consistent, improving $62.6\%$ of the held-out
cases with a signed-rank $p=2.2\times10^{-7}$, whereas over-weighting the structural term
reverses it. Two ablations bound the effect. Adding any third model to the two-model ensemble
degrades it, and classical unsharp masking fails to improve SSIM at any strength (best $0.8765$
against $0.8767$), so the gain reflects learned rather than indiscriminate sharpening. The
result is a cheap, reproducible post-processing stage that improves an already strong
ensemble without any large-scale retraining.

\end{abstract}

\section{Introduction}
\label{sec:intro}
Magnetic resonance imaging (MRI) is central to the diagnosis, treatment planning, and
monitoring of brain tumors, and multi-parametric MRI in particular underpins much of modern
neuro-oncology~\cite{menze2015brats,bakas2017advancing}. Increasingly, these scans are not
read only by clinicians but are also processed by automated analysis pipelines (registration
to anatomical atlases, tissue segmentation, cortical parcellation, and brain
extraction) that were designed and validated on healthy anatomy. When a tumor is present,
such tools become less reliable, because the lesion violates the healthy-anatomy assumptions
on which they depend.

This limitation is difficult to avoid in practice, since a patient is seldom imaged before
disease onset and a healthy anatomical baseline for the same subject is therefore rarely
available~\cite{baid2021rsna}. One way to recover the missing information is to synthesize
it. Given a scan in which the pathological region has been masked out, a generative model can
inpaint the void with plausible healthy tissue while preserving the surrounding anatomy,
producing an image to which healthy-brain pipelines can again be applied. The BraTS
local-synthesis (inpainting) task formalizes this problem and provides a standard benchmark
for it~\cite{kofler2023inpainting}.

The leading 2025 solutions differ from one another by small margins~\cite{kulkarni2025postproc},
which motivates examining whether an existing solution can be improved after synthesis rather
than replaced by a further model. Reports from the strongest recent methods note that their synthesized
tissue is blurry~\cite{zhang2025brats,dai2025context}. The first-place method attributes this
to its mean absolute error (MAE) term, which drives the model toward predicting the mean of
the plausible solutions and so smooths away texture~\cite{zhang2025brats}, and another entry
reports the same effect for a mean squared error (MSE) objective~\cite{dai2025context}. This
blurring bears directly on the ranking. Submissions are ranked by aggregating their ranks on
three metrics, namely the structural similarity index (SSIM), the peak signal-to-noise ratio
(PSNR), and MSE~\cite{labella2024ranking}. Of these, SSIM is the one
that responds most directly to blur, since it measures the structural agreement between a
prediction and the reference and blur removes the fine structure it rewards. Reducing residual
blur is therefore a way to improve the ranking, provided that the pixel-wise metrics are not
degraded in exchange, and applying a sharpening step after synthesis is a natural means to
that end.

To the best of our knowledge, the work most similar to ours is that of Kulkarni et
al.~\cite{kulkarni2025postproc}, from the previous year's challenge, who form an ensemble of
pretrained challenge winners and then apply a broad set of post-processing operations
(classical filtering, histogram matching, and a learned enhancement network) to the ensemble
output. In their evaluation the ensemble is essentially the strongest configuration on its
own, and the additional post-processing, including the learned stage, does not improve upon
it. Our study shares that overall setting, post-processing a deep ensemble of pretrained
winners, but differs in motivation and in scope. Rather than surveying post-processing
operations broadly, we focus on the single effect that the reports above identify, namely
blurriness, and we address it with one learned component, a residual refiner whose training
loss up-weights a structural-similarity term, so that the amount of sharpening applied to the
ensemble can be varied and studied systematically. We assess this choice through controlled
ablations, comprising a sweep of the structural weight, a comparison against classical
unsharp masking, and a variation of the ensemble composition. In contrast to the broad
post-processing of the prior work, we find that an appropriately weighted refiner produces a
small but consistent improvement over the ensemble on the scored metrics.

\section{Related Work}
\label{sec:related}
Image inpainting and, more generally, conditional image synthesis rest on three families of
models. Encoder-decoder regressors built on the U-Net~\cite{ronneberger2015unet} map a masked
image directly to its completion. Conditional generative adversarial networks, exemplified by
pix2pix~\cite{isola2017pix2pix}, add an adversarial objective that encourages sharper and more
realistic outputs. Denoising diffusion probabilistic models~\cite{ho2020ddpm} instead generate
the missing content by iterative refinement, and their latent-space~\cite{rombach2022ldm} and
inpainting-specific~\cite{lugmayr2022repaint} variants are now a common choice for
high-fidelity synthesis. Recent BraTS local-synthesis solutions draw on all three
families~\cite{zhang2025brats,ferreira2025faking,ha2025pseggan}.

The two co-first-place methods of the 2025 edition serve as the base models in this work.
Zhang et al.~\cite{zhang2025brats} use a 3D U-Net trained with
random-masking augmentation, and Ferreira et al.~\cite{ferreira2025faking}
use a fast conditional denoising diffusion model with a hybrid ResNet and
Swin-UNet backbone~\cite{cao2022swinunet}. Both reach high accuracy, and blurring is
discussed in both reports. Zhang et al.\ observe residual blurriness in their inpainted
regions and attribute it to the mean-seeking behaviour of their MAE
term~\cite{zhang2025brats}, while Ferreira et al.\ introduce an MAE term specifically to
reduce the blurring produced by an MSE objective~\cite{ferreira2025faking}. A third entry
reports the same effect for MSE~\cite{dai2025context}.

Both base models therefore already combine a pixel-wise loss with a structural one. Zhang et
al.\ minimize a weighted sum of MAE and SSIM, and Ferreira et al.\ add SSIM and MAE terms to
a diffusion objective. What we examine is not that combination itself, but the effect of
applying it to a separate network placed after a frozen ensemble, and of varying the weight
of the structural term, which neither report examines.

The enhancement stage of Kulkarni et al.~\cite{kulkarni2025postproc}, whose approach is the
closest to ours (Sec.~\ref{sec:intro}), is a lightweight U-Net trained to invert a synthetic
Gaussian blur under an MSE loss, applied to an ensemble formed by pixel and median averaging.
Our refiner differs from it in four respects. First, it is trained on the
actual ensemble outputs ($p_A$, $p_B$, and their disagreement $|p_A-p_B|$) rather than a
synthetic degradation, so the training and inference distributions coincide. Second, it
predicts a residual on the ensemble mean and takes the model-disagreement map as an explicit
input. Third, its loss up-weights SSIM rather than minimizing MSE. Fourth, it post-processes
the 2025 co-winners rather than the 2024 pair. Kulkarni's method was officially ranked while
post-processing pretrained winners, which establishes a precedent for the eligibility of this
class of approach.

Two further 2025 entries are used later in this work as candidate ensemble members.
Local2Global~\cite{saritas2025local2global} is a U-Net with hierarchical attention
mechanisms, and PSegGAN~\cite{ha2025pseggan} is a generative adversarial network conditioned
on a pseudo-segmentation of the surrounding tissue, which its authors report does not surpass
the state of the art on standard voxel-wise metrics. Both are evaluated as third members of
the ensemble in Sec.~\ref{sec:ablation}.

\section{Methods}
\label{sec:methods}
\begin{figure}[tbp]
\centering
\includegraphics[width=\linewidth]{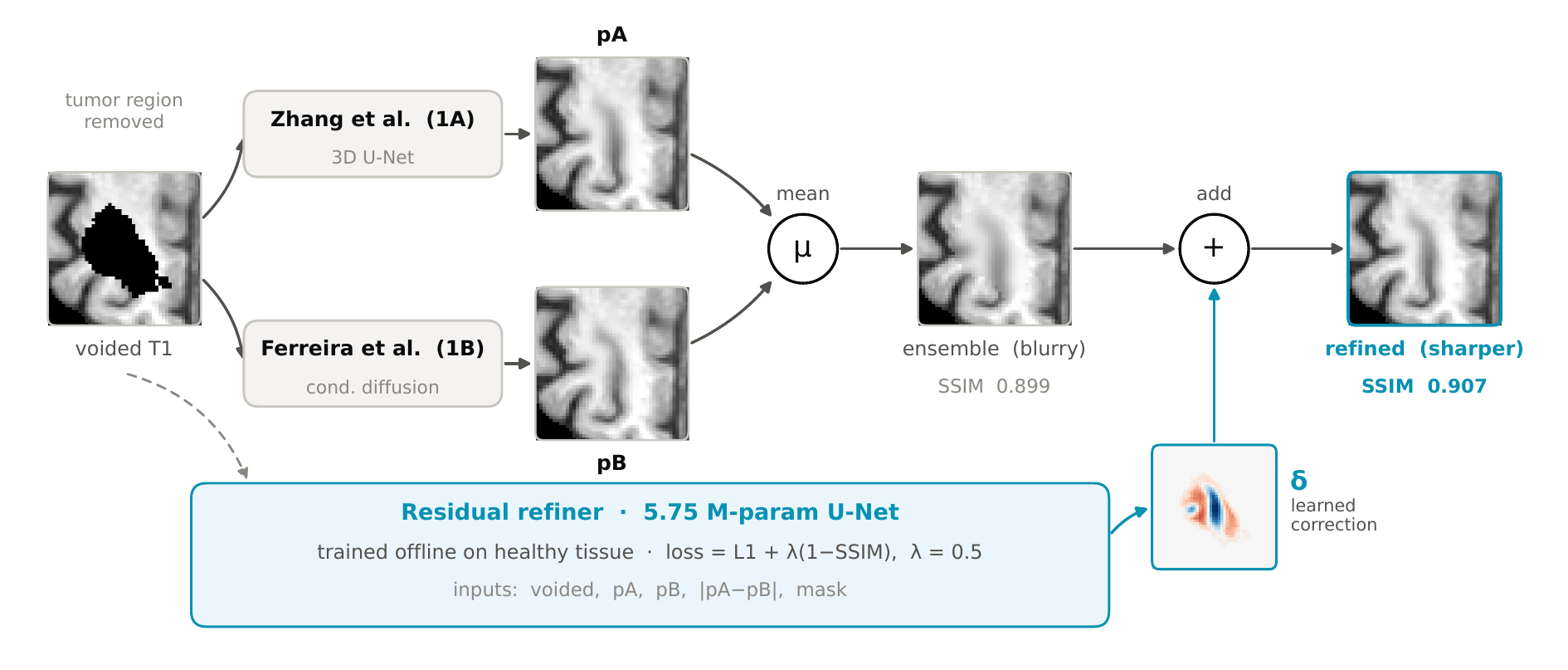}
\caption{Overview of the pipeline. The voided scan is passed to two frozen base models, the
co-first-place entries of the 2025 challenge, 1A, Zhang et al.~\cite{zhang2025brats}, and 1B,
Ferreira et al.~\cite{ferreira2025faking}. Their predictions $p_A$ and $p_B$ are averaged into
the ensemble mean $\bar p$ (Eq.~\ref{eq:ensemble}), which is accurate but over-smoothed. The
residual refiner, trained under Eq.~\ref{eq:loss} with $\lambda=0.5$, receives the voided
scan, both predictions, their disagreement $|p_A-p_B|$ and the void mask, and predicts a
correction $\delta$ added to $\bar p$ (Eq.~\ref{eq:refined}). Neither base model is
fine-tuned. The blur of the ``ensemble (blurry)'' panel is exaggerated for display.}
\label{fig:pipeline}
\end{figure}

Fig.~\ref{fig:pipeline} shows the pipeline as a whole. The base models are used as released,
and the residual refiner is the only component that is trained.

\subsection{Base Models and Deep Ensemble}
\label{sec:bases}
We use the two co-first-place BraTS-2025 algorithms as frozen inference engines, without
fine-tuning. Their predictions are denoted $p_A$ for Zhang et al.~\cite{zhang2025brats} and
$p_B$ for Ferreira et al.~\cite{ferreira2025faking}.\footnote{We run the author-released containers
\texttt{brats25\_inpainting\_ying\_weng} and \texttt{brats25\_inpainting\_faking\_it}, both
from the \texttt{brainles} registry.} The deep
ensemble~\cite{lakshminarayanan2017simple} is their voxel-wise mean,
\begin{equation}
\bar p \;=\; \tfrac{1}{2}\,(p_A + p_B).
\label{eq:ensemble}
\end{equation}
Averaging reduces variance and improves every reported metric over either model alone
(Sec.~\ref{sec:ablation}). It also combines two predictors that both tend toward the
conditional mean, so the ensemble retains the smoothing that both sets of authors report.

\subsection{Residual Refiner}
\label{sec:refiner}
The refiner is a MONAI \texttt{BasicUNet}~\cite{cardoso2022monai} with features
$(32,32,64,128,256,32)$ and $5.75$M parameters. It predicts a residual $\delta$ that is added
to the ensemble mean, so the refined prediction is
\begin{equation}
\hat x \;=\; \bar p + \delta .
\label{eq:refined}
\end{equation}
The network takes five input channels, namely the voided image, $p_A$, $p_B$, the absolute
disagreement $|p_A-p_B|$, and the void mask. The disagreement channel makes the voxel-wise
spread between the two base models available to the network. Intensities are z-scored using
statistics of the voided image, so that the same preprocessing applies during training and at
inference.

\subsection{Training Objective}
\label{sec:loss}
The refiner minimizes
\begin{equation}
\mathcal{L} \;=\; \ell_1(\hat x, x) \;+\; \lambda\,\bigl(1-\mathrm{SSIM}(\hat x, x)\bigr),
\label{eq:loss}
\end{equation}
where $x$ is the ground truth and $\lambda\ge0$ weights the structural term. The loss is
evaluated on the sub-region that the official scorer evaluates.\footnote{The voided region
covers both the tumor and a portion of surrounding healthy tissue, and the scorer evaluates
only the healthy part of it, for which a healthy reference
exists~\cite{kofler2023inpainting,zhang2025brats,dai2025context}. Weighting the loss there
aligns training with evaluation. This is a training-time mask only; inference is unchanged.}
The two terms have different minimizers. The $\ell_1$ term is minimized by the conditional
mean, which favours smooth predictions, whereas $1-\mathrm{SSIM}$ is reduced by preserving
local contrast and structure~\cite{wang2004ssim}. A small $\lambda$ therefore leaves the
objective dominated by $\ell_1$ and reproduces the smoothing of the ensemble, and a larger
$\lambda$ moves the prediction toward sharper local structure. We report a sweep over
$\lambda$ in Sec.~\ref{sec:sweep}.

\subsection{Unsharp Masking Baseline}
\label{sec:unsharp}
As a reference point that involves no learning, we also sharpen the ensemble by unsharp
masking,
\begin{equation}
\hat x_{\mathrm{us}} \;=\; \bar p \;+\; a\,\bigl(\bar p - G_\sigma * \bar p\bigr),
\label{eq:unsharp}
\end{equation}
where $G_\sigma$ is an isotropic 3D Gaussian kernel with standard deviation $\sigma$ in
voxels, $*$ denotes convolution, and $a\ge0$ is the amount. The term $\bar p-G_\sigma*\bar p$
is the difference between the ensemble and a blurred copy of it, so it carries the
high-frequency content of the ensemble, and $a$ sets how much of that content is added back.
Setting $a=0$ recovers the ensemble. Results for a range of $\sigma$ and $a$ are reported in
Sec.~\ref{sec:sharpen}.

\section{Results}
\label{sec:results}

\subsection{Setup}
\label{sec:setup}
Experiments use the BraTS Local-Synthesis dataset~\cite{kofler2023inpainting}, on the
BraTS-GLI~2023 collection~\cite{baid2021rsna,menze2015brats,bakas2017advancing,bakas2017tcgagbm,bakas2017tcgalgg,karargyris2023medperf}.
The challenge provides $1{,}251$ training and $219$ validation cases, each a single
$240\times240\times155$ T1n volume; training cases ship the ground-truth \texttt{t1n} and
sub-masks, while validation cases provide only \texttt{t1n-voided} and the void mask. We
report the official-package metrics SSIM, PSNR, and MSE~\cite{wang2004ssim} on the region the
scorer evaluates, computed both on the official validation leaderboard ($219$ cases) and via
a local reproduction of the official scorer on a held-out set of $219$ training cases. The
latter is a fixed $1{,}032/219$ partition of the training cases by case identifier with seed
$42$, and is distinct from the official validation set, whose ground truth is not released.
The two evaluations are not on the same scale, and the reason is that the base models were
themselves trained on the pool from which our held-out split is drawn, so their predictions
there are better than on data they have not seen. Absolute values are consequently higher on
the held-out split than on the leaderboard, and only the leaderboard is free of this effect
for every component of the pipeline. We also report MAE on the same region, computed after
rescaling both volumes to $[0,1]$ using the ground-truth range inside that region.

The refiner is trained on the $1{,}032$-case portion of the split, from the outputs that the
base models produce for those cases. Training uses $96^3$ patches and
AdamW~\cite{loshchilov2019adamw} with learning rate $10^{-4}$ and weight decay $10^{-5}$, a
batch size of two cases with four patches sampled per case, a $1{,}000$-step linear warmup
followed by cosine decay over $100{,}000$ optimization steps, and bfloat16 precision. Each
refiner trains in $4.8$ hours on a single RTX 3090. At inference the refinement step is a
single pass over the volume and adds $0.36$ s per case on the same hardware (range $0.24$ to
$0.56$ s), on top of running the two base models, which our method leaves unchanged. All refiners share the architecture, data,
schedule, and seed of Sec.~\ref{sec:methods}, differing only in the SSIM weight $\lambda$.

\subsection{Ensemble Composition}
\label{sec:ablation}
Table~\ref{tab:ablation} reports the base models and the ensembles without any refiner. The
two-model mean improves on both individual models, which is the variance reduction expected
of a deep ensemble~\cite{lakshminarayanan2017simple}, and the size of the gain depends on
which models are averaged. Adding a weaker third model
(Local2Global~\cite{saritas2025local2global}, PSegGAN~\cite{ha2025pseggan}, or the
first-place model of the 2024 edition~\cite{zhang2024unet}, which comes from the same group
as $p_A$) moves the mean away from the two strongest predictions and lowers the score. All
three additions degrade the ensemble, and the size of the degradation follows the accuracy of
the model that is added. Evaluated alone on the same split, Local2Global, PSegGAN and the
2024 winner reach SSIM $0.6948$, $0.8141$ and $0.8518$ respectively, and adding them to the
ensemble costs $-0.0423$, $-0.0084$ and $-0.0033$ SSIM. The two-model ensemble is therefore
the configuration that the refiner post-processes in the remainder of this section.

\begin{table}[tbp]
\centering
\caption{Ensemble-composition ablation on the scored region, held-out reproduction, with the
95\% confidence interval on SSIM. The two-model mean improves on either model alone, and
adding any third model degrades it.}
\label{tab:ablation}
\renewcommand{\arraystretch}{1.15}
\begin{tabular}{l cccc}
\toprule
\textbf{Method} & \textbf{SSIM}$\uparrow$ & \textbf{PSNR}$\uparrow$ & \textbf{MSE}$\downarrow$ & \textbf{MAE}$\downarrow$\\
\midrule
Zhang et al.\ ($p_A$)~\cite{zhang2025brats}      & 0.8735 {\scriptsize[.861,.886]} & 24.17 & 0.004204 & 0.03609\\
Ferreira et al.\ ($p_B$)~\cite{ferreira2025faking} & 0.8677 {\scriptsize[.854,.881]} & 23.72 & 0.004823 & 0.03861\\
\textbf{Ensemble ($p_A{+}p_B$)}                & \textbf{0.8767} {\scriptsize[.864,.889]} & \textbf{24.43} & \textbf{0.004020} & \textbf{0.03526}\\
\midrule
\;\;+\,Local2Global~\cite{saritas2025local2global} & 0.8344 {\scriptsize[.819,.849]} & 21.41 & 0.009529 & 0.05619\\
\;\;+\,PSegGAN~\cite{ha2025pseggan}                & 0.8683 {\scriptsize[.855,.881]} & 23.71 & 0.004444 & 0.03935\\
\;\;+\,Zhang et al.\ (2024)~\cite{zhang2024unet}   & 0.8734 {\scriptsize[.861,.886]} & 24.26 & 0.004045 & 0.03601\\
\bottomrule
\end{tabular}
\end{table}

\subsection{Effect of the Structural Weight}
\label{sec:sweep}
Table~\ref{tab:sweep} reports the sweep over $\lambda$ in Eq.~\ref{eq:loss} against the plain
ensemble. At $\lambda=0.1$, where the objective is dominated by $\ell_1$, the refiner improves
MAE over the ensemble and matches its SSIM. At $\lambda=0.5$ it reaches the highest SSIM and
the lowest MAE of the sweep and exceeds the ensemble on SSIM. Beyond that value
the score declines, and at $\lambda=5$ the refiner falls below the ensemble on SSIM. The same
ordering appears on the official validation leaderboard (Table~\ref{tab:official}), which is
not used for training or tuning, where $\lambda=0.5$ is again the highest setting and
$\lambda=5$ is again below the ensemble. The differences between neighbouring settings are
small, as expected when post-processing an ensemble that is already close to the ceiling of
the metric, and the paired comparison below quantifies them per case.

\begin{table}[tbp]
\centering
\caption{Sweep over the structural weight $\lambda$ (Eq.~\ref{eq:loss}) against the plain
ensemble, on a local reproduction of the official scorer (held-out split). Best per column
in \textbf{bold}. SSIM is given with its 95\% confidence interval; these intervals overlap,
and Table~\ref{tab:paired} reports the corresponding paired comparison.}
\label{tab:sweep}
\renewcommand{\arraystretch}{1.15}
\begin{tabular}{l cccc}
\toprule
\textbf{Method} & \textbf{SSIM}$\uparrow$ & \textbf{PSNR}$\uparrow$ & \textbf{MSE}$\downarrow$ & \textbf{MAE}$\downarrow$\\
\midrule
Ensemble ($p_A{+}p_B$)              & 0.8767 {\scriptsize[.864,.889]} & \textbf{24.43} & 0.004020 & 0.03526\\
\midrule
Refiner, $\lambda=0.1$              & 0.8765 {\scriptsize[.864,.889]} & 24.37 & 0.004017 & 0.03498\\
Refiner, $\lambda=0.5$              & \textbf{0.8780} {\scriptsize[.866,.890]} & 24.40 & \textbf{0.004016} & \textbf{0.03491}\\
Refiner, $\lambda=1.0$              & 0.8777 {\scriptsize[.865,.890]} & 24.39 & 0.004031 & 0.03505\\
Refiner, $\lambda=2.0$              & 0.8773 {\scriptsize[.865,.890]} & 24.35 & 0.004061 & 0.03525\\
Refiner, $\lambda=5.0$              & 0.8752 {\scriptsize[.863,.888]} & 24.33 & 0.004076 & 0.03537\\
\bottomrule
\end{tabular}
\end{table}

\begin{table}[tbp]
\centering
\caption{Official validation leaderboard, which is not used for training
or tuning. The sweep again reaches its highest SSIM at $\lambda{=}0.5$, and $\lambda{=}5$ is
again below the plain ensemble, matching the ordering of Table~\ref{tab:sweep} at a different
absolute scale. Best per column in \textbf{bold}.}
\label{tab:official}
\renewcommand{\arraystretch}{1.15}
\begin{tabular}{l cc}
\toprule
\textbf{Method (official)} & \textbf{SSIM}$\uparrow$ & \textbf{PSNR}$\uparrow$\\
\midrule
Ensemble ($p_A{+}p_B$)              & 0.8555 & 25.09\\
\midrule
Refiner, $\lambda=0.1$              & 0.8557 & \textbf{25.11}\\
Refiner, $\lambda=0.5$              & \textbf{0.8572} & 25.10\\
Refiner, $\lambda=1.0$              & 0.8567 & 25.09\\
Refiner, $\lambda=2.0$              & 0.8566 & 25.06\\
Refiner, $\lambda=5.0$              & 0.8539 & 25.05\\
\bottomrule
\end{tabular}
\end{table}

\textbf{Marginal and paired comparisons.} The marginal $95\%$ confidence intervals in
Table~\ref{tab:sweep} overlap almost entirely, and taken alone they would indicate no
difference between the methods. SSIM varies widely from case
to case with the size and content of the scored region, so a marginal interval is dominated
by the variance between cases rather than by the difference between the refiner and the
ensemble. A paired comparison, in which the same case is scored under both pipelines, removes
the between-case variance and measures the per-case difference
$d_i=\mathrm{SSIM}_{\text{refiner}}(i)-\mathrm{SSIM}_{\text{ens}}(i)$ directly.
Table~\ref{tab:paired} reports this comparison. For $\lambda{=}0.5$ the mean per-case
difference is $+0.00127$ SSIM with a bootstrap interval of $[+0.00071,+0.00182]$ that
excludes zero, $62.6\%$ of the $219$ cases improve, and the Wilcoxon signed-rank test gives
$p=2.2\times10^{-7}$. The difference is therefore small in magnitude but consistent in sign
across cases, which is the structure that the overlapping marginal intervals conceal. For
$\lambda{=}0.1$ the same test gives $p=0.064$ with an interval that includes zero, so that
setting is not distinguishable from the ensemble. Both outcomes agree with the leaderboard,
where $\lambda{=}0.5$ is placed above the ensemble and $\lambda{=}0.1$ is level with it. Of
the two evaluations the leaderboard is the external one, since its cases are used neither by
the base models nor by the refiner, while the paired test describes how the difference is
distributed across the cases of the held-out split.

\begin{table}[tbp]
\centering
\caption{Paired comparison on the scored region, per case, on the held-out split. The
interval is a percentile bootstrap on the mean of the per-case difference $d_i$, the win rate
is the fraction of cases with $d_i>0$, and $p$ is a two-sided Wilcoxon signed-rank test. The
$\lambda{=}0.5$ refiner improves on the ensemble in most cases with an interval that excludes
zero, whereas $\lambda{=}0.1$ is not distinguishable from it.}
\label{tab:paired}
\renewcommand{\arraystretch}{1.15}
\setlength{\tabcolsep}{4pt}
\resizebox{\linewidth}{!}{%
\begin{tabular}{l ccccc}
\toprule
\textbf{Comparison} & \textbf{N} & \textbf{mean }$\Delta$\textbf{SSIM} & \textbf{95\% CI} & \textbf{win \%} & $p$ \textbf{(Wilc.)}\\
\midrule
$\lambda{=}0.5$ vs.\ ensemble & 219 & $+0.00127$ & $[+0.00071,\,+0.00182]$ & $62.6$ & $2.2{\times}10^{-7}$\\
$\lambda{=}0.1$ vs.\ ensemble & 219 & $-0.00027$ & $[-0.00112,\,+0.00040]$ & $54.8$ & $0.064$\\
\bottomrule
\end{tabular}}
\end{table}

\subsection{Comparison with Unsharp Masking}
\label{sec:sharpen}
\begin{figure}[tbp]
\centering
\includegraphics[width=\linewidth]{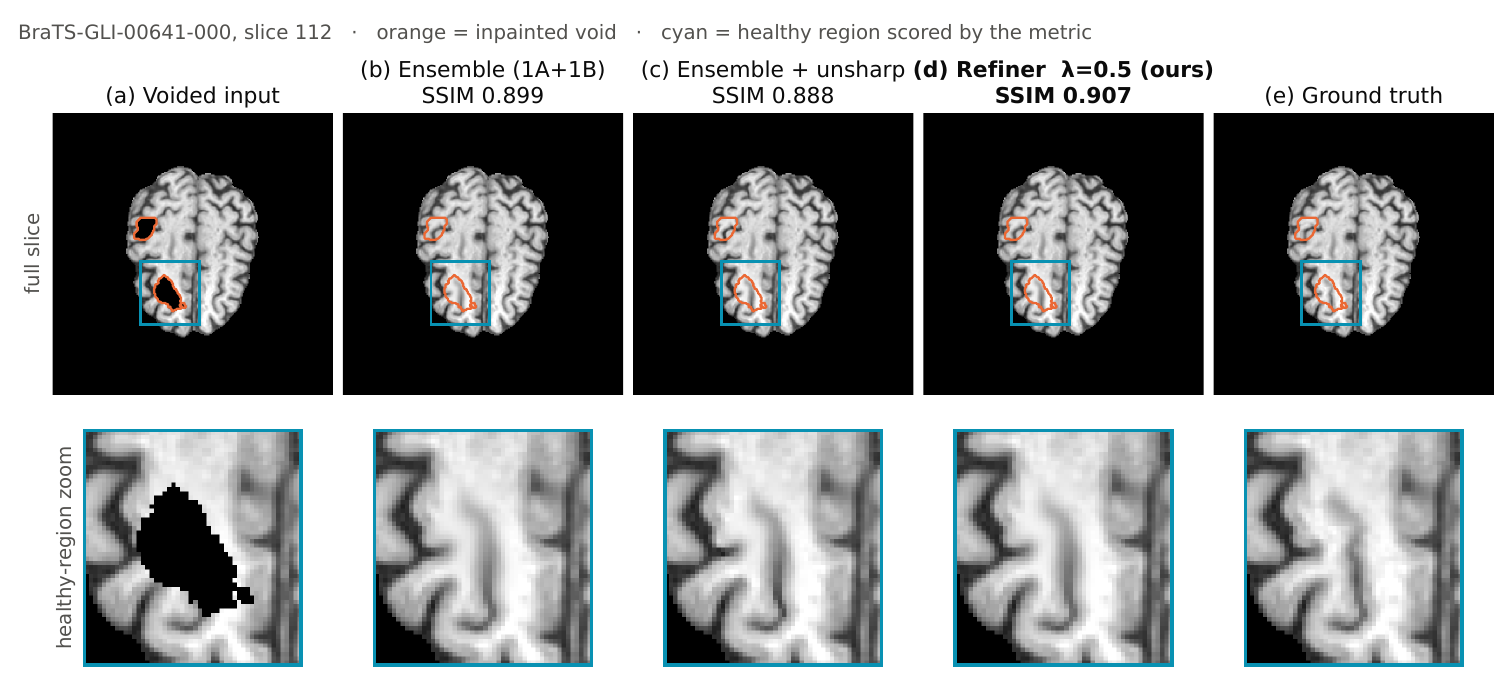}
\caption{Qualitative comparison on a held-out case (BraTS-GLI-00641-000, slice~112):
(a) voided input, (b) the ensemble ($p_A{+}p_B$), (c) the ensemble after unsharp masking with
$\sigma{=}1.0$ and $a{=}1.0$, (d) the $\lambda{=}0.5$ refiner, and (e) ground truth. Orange
outlines the inpainted void; the cyan box marks the healthy region the metric scores,
enlarged in the bottom row. On this case the refiner raises the scored SSIM from $0.899$ to
$0.907$, whereas unsharp masking lowers it to $0.888$, consistent with
Table~\ref{tab:sharpen}.}
\label{fig:qual}
\end{figure}

The refiner adds high-frequency content to the ensemble, which a classical filter also does.
Table~\ref{tab:sharpen} compares the two. We apply unsharp masking (Eq.~\ref{eq:unsharp}) to
the ensemble over a grid of kernel widths $\sigma$ and amounts $a$. No setting improves the
scored SSIM over the plain ensemble, the closest being $0.8765$ against $0.8767$, and at a
fixed kernel width MSE increases monotonically with the amount, from $0.004272$ at $a{=}0.5$
to $0.007139$ at $a{=}2.0$. Unsharp masking amplifies the high-frequency content already
present, including content that does not correspond to the underlying anatomy, so SSIM does
not improve while the pixel-wise error grows, whereas the refiner raises SSIM to $0.8780$
without increasing MSE. The difference therefore lies in which high-frequency content is
added, rather than in the amount of sharpening applied. Fig.~\ref{fig:qual} shows the same
comparison on one case.

\begin{table}[tbp]
\centering
\caption{Unsharp masking of the ensemble (Eq.~\ref{eq:unsharp}) against the refiner, on the
held-out reproduction, where $\sigma$ is the Gaussian kernel width in voxels and $a$ is the
amount. No unsharp setting improves the scored SSIM, and MSE increases monotonically with
$a$. Best per column in \textbf{bold}.}
\label{tab:sharpen}
\renewcommand{\arraystretch}{1.15}
\begin{tabular}{l ccc}
\toprule
\textbf{Configuration} & \textbf{SSIM}$\uparrow$ & \textbf{PSNR}$\uparrow$ & \textbf{MSE}$\downarrow$\\
\midrule
Ensemble (no sharpening)              & 0.8767 & \textbf{24.43} & 0.004020\\
\midrule
Unsharp $\sigma{=}1.0,\,a{=}0.5$      & 0.8765 & 23.99 & 0.004272\\
Unsharp $\sigma{=}0.7,\,a{=}1.0$      & 0.8754 & 23.78 & 0.004411\\
Unsharp $\sigma{=}1.0,\,a{=}1.0$      & 0.8728 & 23.11 & 0.004904\\
Unsharp $\sigma{=}1.5,\,a{=}1.0$      & 0.8672 & 22.02 & 0.006005\\
Unsharp $\sigma{=}1.0,\,a{=}1.5$      & 0.8667 & 22.14 & 0.005876\\
Unsharp $\sigma{=}1.0,\,a{=}2.0$      & 0.8587 & 21.21 & 0.007139\\
\midrule
Refiner, $\lambda{=}0.5$ (learned)    & \textbf{0.8780} & 24.40 & \textbf{0.004016}\\
\bottomrule
\end{tabular}
\end{table}

\section{Discussion}
The results indicate that the weight of the structural term, rather than the capacity of the
refiner, governs whether this form of post-processing improves the score. The $\ell_1$ term
and SSIM are minimized by different predictions, so a refiner trained under an
objective dominated by $\ell_1$ reproduces the smoothing of the ensemble it is meant to
correct, while a refiner trained with too large a structural weight over-sharpens and also
loses score. Zhang et al.~\cite{zhang2025brats} report a related observation from the other
direction, having found a purely structural objective to preserve masked regions poorly and
to introduce artifacts at mask boundaries. Between those two regimes the ordering of the
settings is the same on our held-out reproduction and on the official validation leaderboard.

The comparison with unsharp masking separates two explanations for that improvement. Both the
refiner and the filter add high-frequency content, but only the refiner improves SSIM and only
the filter increases MSE, which is consistent with the refiner adding content that corresponds
to the anatomy it was trained on rather than amplifying what the ensemble already contains.

\textbf{Limitations.} The differences between the ensemble and the refined predictions are
small and their marginal confidence intervals overlap, so the comparison rests on the paired
test of Table~\ref{tab:paired} rather than on the marginal means, and the claim is that the
difference is consistent rather than large. Each configuration is
trained with a single seed, and the differences between neighbouring values of $\lambda$ are
of the order of $10^{-3}$ in SSIM, which need not exceed the variation that retraining with a
different seed would introduce; the leaderboard results, obtained on cases that no part of
the pipeline has seen, are the evidence least affected by this. Absolute scores depend on the
released scorer, which we reproduce locally in order to cross-check the leaderboard. The
refiner is trained for one pair of base models, and its transfer to a different ensemble is
untested.

\section{Conclusion}
Several recent brain-MRI inpainting models report residual blur and attribute it to a
mean-seeking training loss. We post-process an ensemble of two such models with a residual
refiner whose loss combines $\ell_1$ with a structural-similarity term, and report how the
weight of that term affects the result. At a moderate weight the refiner improves SSIM over
the ensemble on both a held-out reproduction of the official scorer and the official
validation leaderboard, with little change in the pixel-wise metrics and without retraining
any base model. The improvement is small but consistent across cases, while the setting
dominated by the $\ell_1$ term is not distinguishable from the ensemble.

\FloatBarrier
\section*{Acknowledgements}
\small Data used in this publication were obtained as part of the Challenge
project through Synapse ID (syn74274097). Experiments used a single NVIDIA
RTX 3090.

\bibliographystyle{splncs04}
\bibliography{refs}

\end{document}